\documentclass[letterpaper]{article} 
\usepackage[draft]{aaai2026}  
\usepackage{times}  
\usepackage{helvet}  
\usepackage{courier}  
\usepackage[hyphens]{url}  
\usepackage{graphicx} 
\usepackage{natbib}  
\usepackage{caption} 
\usepackage{multirow}
\usepackage{multicol}
\usepackage{amssymb}
\usepackage{makecell}
\usepackage[dvipsnames]{xcolor}

\usepackage{algorithm}
\usepackage{algorithmic}

\usepackage{newfloat}
\usepackage{listings}
\DeclareCaptionStyle{ruled}{labelfont=normalfont,labelsep=colon,strut=off} 
\floatstyle{ruled}
\newfloat{listing}{tb}{lst}{}
\floatname{listing}{Listing}
\title{When Words Change the Model: Sensitivity of LLMs for Constraint Programming Modelling}
\author{
    Alessio Pellegrino,\textsuperscript{\rm 1},
    Jacopo Mauro\textsuperscript{\rm 1}
}
\affiliations {
    \textsuperscript{\rm 1}University of Southern Denmark\\
    alessio@imada.sdu.dk, mauro@imada.sdu.dk
}

\usepackage[
     disable,     
    textwidth=.9\marginparwidth,
    colorinlistoftodos,
    prependcaption,
    linecolor=green,backgroundcolor=green!25,bordercolor=green
]{todonotes}

\makeatletter
\if@todonotes@disabled
\else 
\paperwidth=\dimexpr \paperwidth + 5cm\relax
\oddsidemargin=\dimexpr\oddsidemargin + 2cm\relax
\evensidemargin=\dimexpr\evensidemargin + 2cm\relax
\marginparwidth=\dimexpr \marginparwidth + 0.5cm\relax

\fi
\makeatother

\usepackage{amsmath}
\usepackage[capitalize,noabbrev]{cleveref}
\usepackage{nameref}

\newcommand{\vheader}[1]{%
  \makecell[c]{\rotatebox{-90}{#1\hspace{0.8em}}}%
}

\begin{document}

\maketitle

\begin{abstract}

One of the long-standing goals in optimisation and constraint programming is to describe a problem in natural language and automatically obtain an executable, efficient model. Large language models appear to bring this vision closer, showing impressive results in automatically generating models for classical benchmarks. However, much of this apparent success may derive from data contamination rather than genuine reasoning: many standard CP problems are likely included in the training data of these models. To examine this hypothesis, we systematically rephrased and perturbed a set of well-known CSPLib problems to preserve their structure while modifying their context and introducing misleading elements. We then compared the models produced by three representative LLMs across original and modified descriptions. Our qualitative analysis shows that while LLMs can produce syntactically valid and semantically plausible models, their performance drops sharply under contextual and linguistic variation, revealing shallow understanding and sensitivity to wording.
\end{abstract}

\section{Introduction}
\label{sec:introduction}

One of the long-standing goals, the ``holy grail'', of combinatorial optimisation and constraint programming (CP) is to enable users to describe a problem in natural language and automatically obtain an executable, efficient model \cite{bartak1999constraint}.
Such a capability would profoundly democratise optimisation: non-experts could express their needs in everyday terms, and the system would produce a formulation capable of finding solutions under realistic constraints.

This vision has remained elusive, largely because translating an informal description into a formal model requires deep understanding of both the problem semantics and the structure of the target modelling language and the solvers that will execute it.
The advent of large language models (LLMs) has reignited this ambition. 
Their apparent capacity to reason, generalise, and generate structured text suggests that they could serve as universal model generators, bridging natural language and formal optimisation frameworks. 
Recent studies such as \cite{michailidis2025cpbench} and \cite{songllms} have shown that LLMs can indeed produce syntactically valid models for classical CP benchmarks. 
These results have prompted optimism that LLMs may soon automate the knowledge-intensive phases of modelling.

However, we conjecture that a significant portion of these promising results may not stem from genuine reasoning or abstraction abilities, but rather from data contamination. 
Because LLMs are trained on vast amounts of publicly available data, including repositories, tutorials, and benchmark collections, many classical optimisation problems and their correct formulations are likely present in their training corpora. 
Consequently, when evaluated on known benchmarks, the models might reproduce memorised formulations instead of deriving new ones through reasoning.

To investigate this hypothesis, we design a systematic evaluation of LLMs' modelling behaviour under controlled conditions. 
We reformulate well-known CP problems from \textit{CSPLib}, modifying their context and introducing distracting or irrelevant information, while preserving their underlying combinatorial structure. 
This procedure aims to test whether LLMs can recognise structural equivalence beyond surface-level similarity. 
By comparing the models generated from original and modified problem descriptions, we assess the extent to which current LLMs rely on memorisation versus genuine abstraction.

Our study offers qualitative insights into the limits of LLM-driven modelling, highlighting how subtle linguistic cues, context shifts, and distractors affect the correctness and expressiveness of the resulting formulations.

The remainder of the paper is structured as follows. In~\nameref{sec:background} we provide the necessary background and discuss related work. In~\nameref{sec:methodology} we describe our experimental methodology. The results of our analysis are presented in~\nameref{sec:results}, followed in \nameref{sec:discussion} by a broader discussion outlining also the limitations and potential threats to validity of our study. \nameref{sec:conclusions} concludes the paper.
\section{Background and Related Work}\label{sec:background}
\subsection{Constraint Programming}

Constraint Programming \citep{rossi2006handbook} is a paradigm for solving combinatorial problems by expressing them in terms of variables, domains, and constraints. Constraints in CP can take different forms, such as binary constraints, integer constraints or constraints that map relations over multiple variables or arrays of variables. Over the years, CP has been extended with global constraints, which capture commonly occurring substructures (e.g., “all-different,” “cumulative,” “circuit”) and come with specialised searching algorithms. These high-level constraints enable more compact and expressive modelling while also offering more powerful propagation.
There are many languages for writing CP models, of which the most popular are Minizinc \citep{nethercote2007minizinc}, CPMpy \citep{guns2019increasing} and Essence \citep{frisch2007design}. Although all of these languages offer some kind of automatic model reformulation to improve the solving performance, it is still critical to write a performant model to solve the expressed problem efficiently. Even if different problems require different modelling decisions, in general, a good model: (i) has a compact representation to reduce the number of decision variables, (ii) uses global constraints to enhance the propagation, and (iii) uses symmetry-breaking constraints to avoid duplicate solutions \cite{minizinc27Effective}.

\subsection{Large Language Models}
Large Language Models (LLMs) are powerful machine learning models that learn language patterns. LLMs are usually based on deep learning techniques, and they have become popular in many fields due to how they seem to generalise and have good performance on different tasks, even those not strictly related to language. These models usually have billions of parameters and, to avoid overfitting, they are trained on hundreds of billions of textual inputs \cite{hoffmann2022training}. To gather the necessary training data, AI companies often rely on datasets gathered by crawling the whole web. Since these datasets are not public and due to the vast amount of information present in them, a pressing issue in the LLM evaluation process is to understand if there could be any data contamination, that is, if the task on which the LLM is being evaluated on, is present in the training data or not \cite{sainz2023nlp}.
Thanks to their \textit{emergent capabilities} of generalising to tasks beyond purely text-based ones, LLMs have been applied across a wide range of domains, including medical diagnosis \cite{panagoulias2023evaluating}, materials science \cite{jablonka202314}, and physics \cite{barman2025large}.

\subsection{Related Work}

For an overview of different ways of using machine learning techniques to enhance the solution of constraint problems we invite the reader to the comphrehensive survey by \citet{bengio2021machine}. Within the Constraint Programming community, LLMs have been predominantly explored for their potential in the modelling phase, where the goal is to translate a natural language problem description into a formal model written in a constraint modelling language. For example, \citet{penco2025large} presents a framework to generate CP models using LLMs that mitigates the issue of the small amount of code written in constraint modelling languages available as training data. However, they do not investigate the effects of data contamination on their results. \citet{dakle2023ner4opt} and \citet{kaggleDatasetFinetuning} focus of name entity recognition using LLMs to understand the problem and get the information relevant to the modelling phase. This, however, is only a preliminary step and they do not generate directly models.

In particular, several benchmark datasets have recently been introduced to test LLMs' capabilities on modelling constraint problems. CP-Bench \cite{michailidis2025cpbench} provides a collection of several combinatorial problems and supports correctness evaluation across multiple modelling languages, but does not include any checks for data contamination. CPEVAL \cite{songllms} includes 30 problems from CSPLib \cite{gent1999csplib} and presents an initial investigation on the effects of trying to rephrase the problem descriptions making it more schematic but without adding any form of distraction.

Beyond model generation, LLMs have also been used to improve existing CP models. For instance, \citet{voboril2024generating} employs LLMs to generate {streamliners}, which are additional constraints intended to reduce the search space and improve solving efficiency, albeit at the risk of excluding some valid solutions.

\section{Methodology}\label{sec:methodology}
Our main objective is to conduct a qualitative analysis of the constraint programming models generated by large language models (LLMs) and to evaluate their reasoning and modelling capabilities. To this end, we selected the first ten problems from CSPlib~\cite{gent1999csplib} and introduced controlled modifications to their descriptions to reduce the likelihood that an LLM simply recalls a memorised model but still preserving the underlying structure of the problem.

We employed three distinct LLMs: GPT-4~\cite{openaiGPT4} from OpenAI, Claude~4~\cite{anthropicIntroducingClaude} from Anthropic, and DeepSeek-R1~\cite{guo2025deepseek} from DeepSeek. GPT-4 represents one of the most advanced general-purpose LLMs currently available, exhibiting strong performance across a wide range of reasoning, coding, and multilingual tasks.\footnote{Note that we deliberately decided to not use the latest OpenAI version (GPT5) because for our methodology we want to avoid randomness and GPT-5 does not allow to set the temperature to 0 to make the experiments deterministic.} Claude 4, developed by Anthropic, is vastly considered one of the best coding models also thanks to the official claude code tool\footnote{https://claudecode.io/}. DeepSeek-R1, in contrast, is a recently introduced open-weight model explicitly designed for enhanced reasoning. It incorporates a \emph{reinforced reasoning} training paradigm, in which the model generates and leverages internal \textit{thinking} tokens to support intermediate reasoning steps and improve logical consistency. These models were selected as they are state of the art model with different capabilities.
All models were evaluated under identical conditions: the same system prompt was used, all models were accessed via their public APIs, and the temperature was set to 0 to ensure reproducibility. Each model was given a single attempt at producing a solution, with no opportunity for refinement or correction. No in-context examples or demonstrations were provided. 

Considering all problems, we first started by estimating the potential presence of data contamination.
The only definitive way to verify this would be to inspect the LLMs’ training data directly. However, since such data are not publicly available for the considered LLMs, we designed a simple experiment to estimate the degree of prior knowledge each LLM may possess for each problem. We asked each LLM to complete the problem description given only its first half description. A high similarity between the generated and original second halves would suggest a greater likelihood of contamination.

As with the other analyses in this paper, the comparison between the original and generated descriptions was performed manually, and we did not employ any automated similarity metric. In general, we considered two descriptions to be similar if they shared the same objective, phrasing structure, and core conceptual framing. For example, the second half of the Social Golfers problem states: ``\textit{They would like you to come up with a schedule of play for these golfers, to last as many weeks as possible, such that no golfer plays in the same group as any other golfer on more than one occasion.}'' The version generated by Claude 4 was: ``\textit{They want to schedule the golfers into groups over multiple weeks such that every pair of golfers plays together in the same group exactly once, and no pair of golfers plays together more than once.}'' These two descriptions were considered very similar, as they preserved the same objective and structural meaning. In contrast, the original description of the Perfect Square problem states: ``\textit{For a perfect placement problem, all squares have different sizes. The sum of the square surfaces is equal to the surface of the packing square, so that there is no spare capacity.} [...]'' whereas the description generated by GPT-4 was: ``\textit{For a perfect placement problem, a solution is considered valid only if the big square's area is exactly equal to the sum of the areas of the smaller squares.}'' Although both sentences refer to area equality, the generated version omits that all squares must have different sizes, which is essential to the definition. Therefore, we did not consider these descriptions similar.

We then proceed to modify the problem introducing the following two types of changes:
\begin{itemize}
    \item \textbf{Context modification:} We altered the context or background of the problem description. For instance, if the original formulation described an industrial scheduling task, we might reformulate it as a board game scenario. The purpose of this change is to hinder the direct retrieval of a pre-existing model from memory. When an LLM still produces a correct model under these conditions, it suggests a higher level of reasoning ability.
    \item \textbf{Distraction modification:} We introduced additional sentences unrelated to the actual problem or potentially misleading regarding its objective. For example, a satisfaction problem might be disguised as an optimisation one, or vice versa. These modifications introduce noise into the problem description, reflecting more realistic, imperfect input conditions. We argue that the ability to distinguish between essential and irrelevant information is fundamental if LLMs are to be used as autonomous or semi-autonomous modellers.
\end{itemize}

As an illustrative example consider the \textit{Sudoku} problem, a famous puzzles well studied in the CP community~\cite{DBLP:journals/tplp/DemoenB14}.
Sudoku is a logic-based number placement puzzle played on a 9×9 grid divided into nine 3×3 subgrids.
The goal is to fill the grid with digits 1 to 9 so that each row, each column, and each 3×3 subgrid contains every digit exactly once.
For the context modification, we reformulated the problem as arranging not digits but queens with different adornments on a \(9 \times 9\) chessboard: ``\textit{My game is inspired by chess with similar pieces but it has different rules. It involves positioning n queens with different garments on a n by n chess board where n = 9. The queens can only attack queens with the same garment}''.
As a distraction instead we change the formulation of the objective, stating "\textit{place queens such to maximise the number of queens garments on the board while, at the same time, avoid attacks between queens}".

Each problem was first rewritten with only a context modification, and then with both a context and a distraction modification. The aim was to maintain equivalence with the original problem, such that a valid model for the modified version would also be valid for the original one, and vice versa. As a target platform to solve the problem, we require the model to written in the MiniZinc~\cite{nethercote2007minizinc} language to be run with its standard toolchain.

We evaluated the modelling capabilities of the three LLMs on the original and modified problem descriptions (context only, and context \& distraction). The quality of the generated models was assessed according to two metrics:

\begin{itemize}
    \item \textbf{correct:} whether the generated model represents the correct problem. We considered a model correct if it clearly attempts to solve the intended problem, even if some constraints were expressed incorrectly; such cases were marked as correct but not running.
    \item \textbf{run:} whether the generated model was correct and executed successfully in Minizinc without requiring major manual modifications and using the default MiniZinc IDE settings. Minor adjustments, such as adding missing imports for global constraints, renaming variables, or correcting custom output, were considered acceptable as these could be easily fixed by a user looking at the MiniZinc compiling errors or its output format.
\end{itemize}

In addition to the two quantitative metrics and the contamination check, we also performed a qualitative, manual analysis of the generated models aimed to understand whether different formulations lead to distinct model representations, and more broadly, whether exposure to specific problem descriptions during training may influence the LLM’s modelling performance.

For reproducibility purposes, all prompts, modified problem statements, and generated model outputs used in this study are available at \url{https://github.com/SeppiaBrilla/LLM-cp-eval}, enabling independent verification and follow-up experimentation.

\section{Results}\label{sec:results}
In this section we present the results of our evaluation of the three LLMs. We first discuss the quantitative outcomes based on the two defined metrics, followed by a qualitative analysis of the generated models. In addition to the 10 CSPlib problems chosen as benchmarks, we also provide the data also for the Sudoku problem chosen as an illustrative example in the previous section.

\subsection{Quantitative Results}

As far as the data contamination is concerned, the \textit{Completion} column in \Cref{tab:summary} presents the results of the completion experiment designed to estimate prior knowledge of each LLM regarding the selected problems. The results suggest that GPT-4 and Claude 4 may have prior knowledge of, respectively, four and five problems, as evidenced by their ability to accurately complete four out of ten problem descriptions. Likely, these models were extensively considered when training both of these LLMs. DeepSeek-R1 appears to have even a higher likelihood of contamination, correctly completing six out of ten problems; however, at least one of these models (\textit{Vessel Loading}) may have occurred by chance, as the generated continuation was generic and followed a typical satisfaction problem template.

We assume that when a model produces a description closely matching the original, it likely indicates data contamination. Moreover, if contamination is present, we expect the resulting MiniZinc model to be of higher quality.

\begin{table*}[]
    \small
\resizebox{\linewidth}{!}{\begin{tabular}{|c|c|c|c||c||c|c|c||c|c|c||c|c|c|}
 \hline
 \multirow{2}{*}{Problem} & \multicolumn{3}{|c||}{Completion} & \multirow{2}{*}{Metrics} & \multicolumn{3}{|c||}{Original} & \multicolumn{3}{c||}{Context} & \multicolumn{3}{c|}{\makecell{Context \\ \& \\ Distraction}} \\
 \cline{2-14}
 &  \rule{0pt}{2.2ex} \vheader{Claude}
& \vheader{GPT}
& \vheader{R1}
&
& \vheader{Claude}
& \vheader{GPT}
& \vheader{R1}
& \vheader{Claude}
& \vheader{GPT}
& \vheader{R1}
& \vheader{Claude}
& \vheader{GPT}
& \vheader{R1} \\
 \Xhline{1.2pt}

\multirow{2}{*}{Sudoku}
& \multirow{2}{*}{\textcolor{ForestGreen}{\checkmark}} & \multirow{2}{*}{\textcolor{ForestGreen}{\checkmark}} & \multirow{2}{*}{\textcolor{ForestGreen}{\checkmark}} & Correct & \textcolor{ForestGreen}{\checkmark}&\textcolor{ForestGreen}{\checkmark}&\textcolor{ForestGreen}{\checkmark}&\textcolor{ForestGreen}{\checkmark}&\textcolor{ForestGreen}{\checkmark}&\textcolor{ForestGreen}{\checkmark}&\textcolor{red}{$\times$}&\textcolor{red}{$\times$}&\textcolor{red}{$\times$} \\
\cline{5-14}
& & & & Run & \textcolor{ForestGreen}{\checkmark}&\textcolor{ForestGreen}{\checkmark}&\textcolor{ForestGreen}{\checkmark}&\textcolor{ForestGreen}{\checkmark}&\textcolor{ForestGreen}{\checkmark}&\textcolor{ForestGreen}{\checkmark}&\textcolor{red}{$\times$}&\textcolor{red}{$\times$}&\textcolor{red}{$\times$} \\
\cline{5-14}
\Xhline{1.2pt}

\multirow{2}{*}{Car Sequencing}
& \multirow{2}{*}{\textcolor{ForestGreen}{\checkmark}} & \multirow{2}{*}{\textcolor{ForestGreen}{\checkmark}} & \multirow{2}{*}{\textcolor{ForestGreen}{\checkmark}} & Correct & \textcolor{ForestGreen}{\checkmark}&\textcolor{ForestGreen}{\checkmark}&\textcolor{ForestGreen}{\checkmark}&\textcolor{red}{$\times$}&\textcolor{red}{$\times$}&\textcolor{red}{$\times$}&\textcolor{red}{$\times$}&\textcolor{red}{$\times$}&\textcolor{red}{$\times$} \\
\cline{5-14}
& & & & Run & \textcolor{ForestGreen}{\checkmark}&\textcolor{ForestGreen}{\checkmark}&\textcolor{ForestGreen}{\checkmark}&\textcolor{red}{$\times$}&\textcolor{red}{$\times$}&\textcolor{red}{$\times$}&\textcolor{red}{$\times$}&\textcolor{red}{$\times$}&\textcolor{red}{$\times$} \\
\cline{5-14}
\Xhline{1.2pt}

\multirow{2}{*}{Template Design}
& \multirow{2}{*}{\textcolor{red}{$\times$}} & \multirow{2}{*}{\textcolor{red}{$\times$}} & \multirow{2}{*}{\textcolor{red}{$\times$}} & Correct & \textcolor{ForestGreen}{\checkmark}&\textcolor{ForestGreen}{\checkmark}&\textcolor{ForestGreen}{\checkmark}&\textcolor{ForestGreen}{\checkmark}&\textcolor{ForestGreen}{\checkmark}&\textcolor{ForestGreen}{\checkmark}&\textcolor{red}{$\times$}&\textcolor{red}{$\times$}&\textcolor{ForestGreen}{\checkmark} \\
\cline{5-14}
& & & & Run & \textcolor{ForestGreen}{\checkmark}&\textcolor{red}{$\times$}&\textcolor{ForestGreen}{\checkmark}&\textcolor{ForestGreen}{\checkmark}&\textcolor{red}{$\times$}&\textcolor{ForestGreen}{\checkmark}&\textcolor{red}{$\times$}&\textcolor{red}{$\times$}&\textcolor{ForestGreen}{\checkmark} \\
\cline{5-14}
\Xhline{1.2pt}

\multirow{2}{*}{Quasi Group Existence}
& \multirow{2}{*}{\textcolor{red}{$\times$}} & \multirow{2}{*}{\textcolor{red}{$\times$}} & \multirow{2}{*}{\textcolor{red}{$\times$}} & Correct & \textcolor{ForestGreen}{\checkmark}&\textcolor{ForestGreen}{\checkmark}&\textcolor{ForestGreen}{\checkmark}&\textcolor{ForestGreen}{\checkmark}&\textcolor{ForestGreen}{\checkmark}&\textcolor{ForestGreen}{\checkmark}&\textcolor{ForestGreen}{\checkmark}&\textcolor{ForestGreen}{\checkmark}&\textcolor{ForestGreen}{\checkmark} \\
\cline{5-14}
& & & & Run & \textcolor{red}{$\times$}&\textcolor{red}{$\times$}&\textcolor{red}{$\times$}&\textcolor{red}{$\times$}&\textcolor{red}{$\times$}&\textcolor{ForestGreen}{\checkmark}&\textcolor{ForestGreen}{\checkmark}&\textcolor{red}{$\times$}&\textcolor{ForestGreen}{\checkmark} \\
\cline{5-14}
\Xhline{1.2pt}

\multirow{2}{*}{Secret Shopper}
& \multirow{2}{*}{\textcolor{red}{$\times$}} & \multirow{2}{*}{\textcolor{red}{$\times$}} & \multirow{2}{*}{\textcolor{red}{$\times$}} & Correct & \textcolor{ForestGreen}{\checkmark}&\textcolor{red}{$\times$}&\textcolor{ForestGreen}{\checkmark}&\textcolor{red}{$\times$}&\textcolor{red}{$\times$}&\textcolor{red}{$\times$}&\textcolor{red}{$\times$}&\textcolor{red}{$\times$}&\textcolor{red}{$\times$} \\
\cline{5-14}
& & & & Run & \textcolor{red}{$\times$}&\textcolor{red}{$\times$}&\textcolor{ForestGreen}{\checkmark}&\textcolor{red}{$\times$}&\textcolor{red}{$\times$}&\textcolor{red}{$\times$}&\textcolor{red}{$\times$}&\textcolor{red}{$\times$}&\textcolor{red}{$\times$} \\
\cline{5-14}
\Xhline{1.2pt}

\multirow{2}{*}{\makecell{Low Autocorrelation\\Binary Sequences}}
& \multirow{2}{*}{\textcolor{ForestGreen}{\checkmark}} & \multirow{2}{*}{\textcolor{red}{$\times$}} & \multirow{2}{*}{\textcolor{red}{$\times$}} & Correct & \textcolor{ForestGreen}{\checkmark}&\textcolor{ForestGreen}{\checkmark}&\textcolor{ForestGreen}{\checkmark}&\textcolor{ForestGreen}{\checkmark}&\textcolor{ForestGreen}{\checkmark}&\textcolor{ForestGreen}{\checkmark}&\textcolor{red}{$\times$}&\textcolor{ForestGreen}{\checkmark}&\textcolor{ForestGreen}{\checkmark} \\
\cline{5-14}
& & & & Run & \textcolor{ForestGreen}{\checkmark}&\textcolor{red}{$\times$}&\textcolor{red}{$\times$}&\textcolor{ForestGreen}{\checkmark}&\textcolor{red}{$\times$}&\textcolor{ForestGreen}{\checkmark}&\textcolor{red}{$\times$}&\textcolor{red}{$\times$}&\textcolor{red}{$\times$} \\
\cline{5-14}
\Xhline{1.2pt}

\multirow{2}{*}{Golomb Ruler}
& \multirow{2}{*}{\textcolor{ForestGreen}{\checkmark}} & \multirow{2}{*}{\textcolor{ForestGreen}{\checkmark}} & \multirow{2}{*}{\textcolor{ForestGreen}{\checkmark}} & Correct & \textcolor{ForestGreen}{\checkmark}&\textcolor{ForestGreen}{\checkmark}&\textcolor{ForestGreen}{\checkmark}&\textcolor{ForestGreen}{\checkmark}&\textcolor{red}{$\times$}&\textcolor{ForestGreen}{\checkmark}&\textcolor{red}{$\times$}&\textcolor{red}{$\times$}&\textcolor{ForestGreen}{\checkmark} \\
\cline{5-14}
& & & & Run & \textcolor{red}{$\times$}&\textcolor{red}{$\times$}&\textcolor{ForestGreen}{\checkmark}&\textcolor{ForestGreen}{\checkmark}&\textcolor{red}{$\times$}&\textcolor{ForestGreen}{\checkmark}&\textcolor{red}{$\times$}&\textcolor{red}{$\times$}&\textcolor{ForestGreen}{\checkmark} \\
\cline{5-14}
\Xhline{1.2pt}

\multirow{2}{*}{All-Interval Series}
& \multirow{2}{*}{\textcolor{red}{$\times$}} & \multirow{2}{*}{\textcolor{red}{$\times$}} & \multirow{2}{*}{\textcolor{red}{$\times$}} & Correct & \textcolor{ForestGreen}{\checkmark}&\textcolor{ForestGreen}{\checkmark}&\textcolor{ForestGreen}{\checkmark}&\textcolor{ForestGreen}{\checkmark}&\textcolor{ForestGreen}{\checkmark}&\textcolor{ForestGreen}{\checkmark}&\textcolor{ForestGreen}{\checkmark}&\textcolor{ForestGreen}{\checkmark}&\textcolor{ForestGreen}{\checkmark} \\
\cline{5-14}
& & & & Run & \textcolor{ForestGreen}{\checkmark}&\textcolor{ForestGreen}{\checkmark}&\textcolor{ForestGreen}{\checkmark}&\textcolor{ForestGreen}{\checkmark}&\textcolor{red}{$\times$}&\textcolor{red}{$\times$}&\textcolor{ForestGreen}{\checkmark}&\textcolor{ForestGreen}{\checkmark}&\textcolor{ForestGreen}{\checkmark} \\
\cline{5-14}
\Xhline{1.2pt}

\multirow{2}{*}{Vessel Loading}
& \multirow{2}{*}{\textcolor{red}{$\times$}} & \multirow{2}{*}{\textcolor{red}{$\times$}} & \multirow{2}{*}{\textcolor{ForestGreen}{\checkmark}} & Correct & \textcolor{ForestGreen}{\checkmark}&\textcolor{ForestGreen}{\checkmark}&\textcolor{ForestGreen}{\checkmark}&\textcolor{ForestGreen}{\checkmark}&\textcolor{red}{$\times$}&\textcolor{ForestGreen}{\checkmark}&\textcolor{red}{$\times$}&\textcolor{red}{$\times$}&\textcolor{red}{$\times$} \\
\cline{5-14}
& & & & Run & \textcolor{ForestGreen}{\checkmark}&\textcolor{ForestGreen}{\checkmark}&\textcolor{red}{$\times$}&\textcolor{ForestGreen}{\checkmark}&\textcolor{red}{$\times$}&\textcolor{ForestGreen}{\checkmark}&\textcolor{red}{$\times$}&\textcolor{red}{$\times$}&\textcolor{red}{$\times$} \\
\cline{5-14}
\Xhline{1.2pt}

\multirow{2}{*}{Perfect Square}
& \multirow{2}{*}{\textcolor{red}{$\times$}} & \multirow{2}{*}{\textcolor{red}{$\times$}} & \multirow{2}{*}{\textcolor{ForestGreen}{\checkmark}} & Correct & \textcolor{ForestGreen}{\checkmark}&\textcolor{ForestGreen}{\checkmark}&\textcolor{ForestGreen}{\checkmark}&\textcolor{red}{$\times$}&\textcolor{ForestGreen}{\checkmark}&\textcolor{ForestGreen}{\checkmark}&\textcolor{red}{$\times$}&\textcolor{red}{$\times$}&\textcolor{red}{$\times$} \\
\cline{5-14}
& & & & Run & \textcolor{red}{$\times$}&\textcolor{red}{$\times$}&\textcolor{red}{$\times$}&\textcolor{red}{$\times$}&\textcolor{ForestGreen}{\checkmark}&\textcolor{red}{$\times$}&\textcolor{red}{$\times$}&\textcolor{red}{$\times$}&\textcolor{red}{$\times$} \\
\cline{5-14}
\Xhline{1.2pt}

\multirow{2}{*}{Social Golfers}
& \multirow{2}{*}{\textcolor{ForestGreen}{\checkmark}} & \multirow{2}{*}{\textcolor{ForestGreen}{\checkmark}} & \multirow{2}{*}{\textcolor{ForestGreen}{\checkmark}} & Correct & \textcolor{ForestGreen}{\checkmark}&\textcolor{red}{$\times$}&\textcolor{red}{$\times$}&\textcolor{ForestGreen}{\checkmark}&\textcolor{red}{$\times$}&\textcolor{red}{$\times$}&\textcolor{red}{$\times$}&\textcolor{red}{$\times$}&\textcolor{red}{$\times$} \\
\cline{5-14}
& & & & Run & \textcolor{red}{$\times$}&\textcolor{red}{$\times$}&\textcolor{red}{$\times$}&\textcolor{ForestGreen}{\checkmark}&\textcolor{red}{$\times$}&\textcolor{red}{$\times$}&\textcolor{red}{$\times$}&\textcolor{red}{$\times$}&\textcolor{red}{$\times$} \\
\cline{5-14}
\Xhline{1.2pt}

\multirow{2}{*}{Total}
& \multirow{2}{*}{5} & \multirow{2}{*}{4} & \multirow{2}{*}{6} & Correct & 11 & 9 & 10 & 8 & 6 & 8 & 2 & 3 & 5 \\
\cline{5-14}
& & & & Run & 6 & 4 & 6 & 7 & 2 & 6 & 2 & 1 & 4 \\
\cline{5-14}
\hline

\end{tabular}}
\caption{Results of our experiments on each LLM. The Completion column shows whether a certain LLM correctly finished the description of a given problem (\textcolor{ForestGreen}{\checkmark} when correct and \textcolor{red}{$\times$} when not). The Original column shows, for each problem and LLM, the results on our two metrics for the original description, Context for the context modified description and Context \& Distraction for the context modification with an additional distractor (\textcolor{ForestGreen}{\checkmark} when correct and \textcolor{red}{$\times$} when not).}
\label{tab:summary}
\end{table*}

\Cref{tab:summary} also summarises the performance of the three LLMs across all problems. For each problem and each version of the description (original, context-modified, and context \& distraction), we report the outcome according to the two evaluation metrics.

Overall, we observe that all three LLMs perform strongly on the original descriptions, particularly on the \textit{correct} metric, with Claude 4 achieving a perfect score.

When evaluating the context-modified descriptions, performance decreases across all models. The best result in the \textit{correct} metric is 8 out of 11 correct models (considering also Sudoku), achieved by both Claude 4 and DeepSeek-R1. We also observe a slight decline in the \textit{run} metric, though Claude 4 is not affected.

When context and distraction changes are introduced, performance dropped significantly across all metrics. The largest decline occurs in the \textit{correct} metric: none of the LLMs produced correct formulations for even half of the problems, with DeepSeek-R1 coming closest at 5 out of 11.

As we further elaborate later, LLMs seem particularly reliable when the problem description includes an explicit mathematical formulation (e.g., \emph{Quasi Group Existence}, \emph{All-Interval Series}). In such cases, the models rarely falls for the distraction, indicating that clearly stated formal structure strongly anchors the modelling process.


\subsection{Qualitative Results}

We now discuss in greater depth each problem, its modifications, and the modelling decisions made by the LLMs.

In this analysis, when discussing individual constraints or variables, we refer to their intended purpose rather than whether the final model successfully captures that intention. If a model fails to express a constraint in the intended manner, this is already reflected in the run metric for that model.

\subsubsection{Sudoku}

This problem, unlike the others in our benchmark, was not sourced from CSPlib, and we added it as an illustrative example due to its simplicity and widespread recognition. Due to this popularity, we also conjectured that this problem was very likely used in the training of the LLMS.

For the context modification, as detailed in \nameref{sec:methodology}, we reformulated the problem as arranging queens with different adornments on a \(9 \times 9\) chessboard. This change in context did not affect the performance of the LLMs. GPT-4 explicitly recognised the formulation as a variation of the Sudoku problem, and all models produced equivalent solutions, differing only in variable naming, from those generated for the original version.

The results changed, however, when we introduced a distractor. The structure of the problem remained identical, except for rephrasing the objective adding in particular the keyword ``maximize''. This modification led the LLMs to incorrectly model the task as an optimisation problem, despite the rest of the description imply that it was feasible to place only one queen of a given type in each cell without conflict. We conjecture here that the presence of the word ``maximize'' triggered the LLMs to interpret the task as an optimisation problem, leading them to overlook the underlying satisfaction nature of the original Sudoku formulation.

\subsubsection{Car Sequencing}
involves scheduling the production of a set of cars, each requiring the installation of different optional features. Some options are mutually incompatible, and each must be installed at a specific workstation with limited capacity that cannot be exceeded.
We reformulated the problem by changing its context from car production to the assembly of knapsacks equipped with various accessories. Interestingly, this seemingly minor contextual change caused all evaluated LLMs to fail in generating a correct model. The primary difficulty arose from misinterpreting the requirement to produce a specified number of knapsacks for each type, as all models instead treated the demand as a single global quantity. In other words, instead of having an array \(D\) of demands where \(D_i\) was the required number of knapsacks of type \(i\), there was a single demand variable for all knapsack types. Notably, none of the models made this error when working with the original car production formulation. We believe this could be an effect of the LLM already knowing the car sequencing problem, as the two descriptions referring to this constraint are very similar. The original states: "\textit{A number of cars are to be produced; they are not identical, because different options are available as variants on the basic model.}" While the modified says: "\textit{A company must make a certain number of different knapsacks of different sizes and with different accessories.}" 

When we introduced the distractor, which requested the maximisation of the number of accessories attached to each knapsack (``\textit{A company must make a certain number of different knapsacks of various sizes and with different accessories so that they can maximise the amount of items each knapsack can contain with respect to its size.}''), all three LLMs reformulated the problem as a maximisation task. This misinterpretation immediately rendered all resulting models incorrect.

Interestingly, none of the LLMs produced a valid schedule in either of the modified models. We hypothesise that this issue may stem from the fact that the term ``schedule'' was never explicitly mentioned in the modified problem descriptions. Furthermore, both Claude 4 and DeepSeek-R1 represented the capacity constraint as a percentage, but only in both the modified models, and not in the original ones, despite the explicit use of the word ``percentage'' in both cases. We speculate that this behaviour may be influenced by the fact that most publicly available formulations of the Car Sequencing Problem do not model capacity constraints as percentages.

\subsubsection{Template Design}
This problem involves determining the number of templates required to print a given set of boxes. Each template can accommodate only a limited number of boxes, and the objective is to satisfy the overall demand while minimising the total number of templates used.

We reformulated this problem through a context modification by replacing boxes with advertisement panels. Due to the minimal nature of this transformation, we observed only minor differences between the original and the context-modified models.

However, when introducing a distractor, the situation changes noticeably. We added the following sentence to the problem description: ``\textit{Due to occasional misalignment in the cutting process, some panel arrangements may not be feasible.}'' Although the statement does not explicitly appear in the objective, both Claude 4 and GPT-4 incorporated additional constraints to account for potential misalignment issues. The only exception was DeepSeek-R1, which justified its decision 
by stating: ``\textit{The problem mentions that some arrangements may not be feasible due to misalignment, but no specific constraints are provided.}''

\subsubsection{Quasi Group Existence}

models a Latin square with additional constraints: we want to know if it’s possible to fill in an \(n \times n\) table with \(n\) symbols so that: i) each symbol occurs once per row and column (the Latin square condition), and ii) any extra algebraic constraints are satisfied. In particular, we define a new operation denoted by ``\(*_{321}\)'' such that \(a *_{321} b = c\) if and only if \(c * b = a\). A quasigroup of order \(m\) exists if, for all \(a, b, c, d\), the conditions \(a * b = c\), \(a * b = c * d\), and \(a *_{321} b = c *_{321} d\) together imply that \(a = c\) and \(b = d\).

We reformulated the problem through a context modification by mapping it to a delivery-drone scheduling scenario, where each row represents a drone, each column corresponds to a delivery zone, and the entry value indicates the timestep of delivery. For instance, if drone \(i\) has a value of 4 in zone \(j\), this means that drone \(i\) delivers to zone \(j\) at timestep 4. The performance of all three LLMs remained essentially unchanged compared to the original formulation. Interestingly, all models recognised the structural similarity to a Latin square, even though this connection was not explicitly stated in the context-modified description.

A similar pattern was observed when introducing the distractor, which mentioned the need to keep workloads balanced. All three LLMs continued to focus solely on the Latin square structure and disregarded the distractor information.

The most notable observation for this problem is that each LLM represented the additional constraint on the Latin square differently across all formulations. This suggests that the models likely lacked a single dominant example or canonical formulation of this specific constraint in their training data.

\subsubsection{Secret Shopper}
involves scheduling a series of visits by secret shoppers to workers across different shops, ensuring that each worker is visited four times by distinct shoppers. Additional constraints apply to these visits. Among the evaluated models, only GPT-4 failed to correctly represent the original problem, as it omitted the constraint requiring each worker to be visited by different groups of shoppers.

The modified version of the problem introduced two small changes: the workers were replaced with law enforcement officers, and some redundant numerical details were removed from the description. Despite the seemingly minor adjustments, all three LLMs failed to correctly model the modified context. The models failed by omitting the constraint requiring that each worker be visited by different officers.
The removal of the redundant information that included explicit numerical calculations that likely facilitated correct modelling may explain the observed decline in performance.

We subsequently introduced a workload-balance distractor: ``\textit{Each agent has a personal workload limit to avoid burnout. Ideally, visits should be evenly spaced among agents.}'' This statement was not reiterated in the objective description. Although all three LLMs produced incorrect models for this distractor variant, only Claude 4 and GPT-4 incorporated the workload balance constraint, while DeepSeek-R1 failed for the same reason as in the context-modified version.

\subsubsection{Low Autocorrelation Binary Sequence}
requires designing a binary sequence that minimises autocorrelation between its elements. Each bit can take the value \(-1\) or \(+1\), and the \(k\)-th autocorrelation is defined as
\[
\sum_{i=0}^{n-k-1} S_i \times S_{i+k},
\]
where \(S_i\) denotes the value of the \(i\)-th bit in the sequence. The objective is to minimise the sum of the squares of these autocorrelations.

This problem was relatively easy for all models to represent correctly, primarily because the objective function was explicitly stated in the problem description. We then applied a context modification by reformulating the task as a music composition problem, where the goal was to minimise the repetitiveness of a rhythmic pattern. Since the explicit mathematical formulation was preserved, this change in context did not negatively affect model performance.

The distractor we introduced required the model to consider the influence of consecutive beats of the same type: ``\textit{Since the rhythm will be played on a percussion instrument where consecutive beats of the same type tend to reinforce each other, producing slightly louder or softer passages depending on the sequence, the composer must also take this behaviour into account [...]}'' Despite this addition, both GPT-4 and DeepSeek-R1 continued to model the problem correctly by adhering to the explicit formula. Claude 4, however, introduced an unnecessary constraint enforcing a balance between positive and negative beats, which rendered its model incorrect.

\subsubsection{Golomb Ruler}
can be defined as a set of \(m\) integers \(0 = a_1 < a_2 < \dots < a_m\) such that the \(\frac{m(m - 1)}{2}\) differences \(a_j - a_i\), for \(1 \leq i < j \leq m\), are all distinct. The objective is to find an optimal (i.e., minimum-length) or near-optimal ruler.

We applied a context modification by reframing the problem as the placement of a set of signal beacons along a straight path. Among the evaluated models, only GPT-4 produced an incorrect formulation, as it enforced distinctness only among consecutive distances rather than all pairwise distances. Interestingly, this was one of the only two instances in which an LLM explicitly recognised the similarity with the original formulation, DeepSeek-R1 stated: ``\textit{The goal is to minimize the position of the last beacon to reduce cost, which aligns with the Golomb ruler problem.}''

To introduce a distractor, we mentioned that different beacons could have varying ranges and costs. Both GPT-4 and Claude 4 incorporated an additional constraint to model the cost of beacons, whereas DeepSeek-R1 maintained its previous formulation, again referencing its correspondence to the original Golomb ruler problem.

\subsubsection{All-interval Series}
requires, given the twelve standard pitch classes, finding a series in which each pitch class occurs exactly once and the musical intervals between consecutive notes cover the complete set of intervals from the minor second (1 semitone) to the major seventh (11 semitones). In other words, for every possible interval, there must exist a pair of adjacent pitch classes in the sequence separated by that interval.

This task was straightforward for all evaluated LLMs, likely because the problem description was highly detailed. Even after recontextualising the problem as one of constructing a specific DNA sequence, all models successfully generated correct formulations.

The same outcome was observed when introducing a distractor describing temperature constraints between DNA samples: ``\textit{Some DNA samples are sensitive to temperature fluctuations. The lab technician wants to avoid large jumps in base-pair counts between consecutive experiments because it could require recalibrating the equipment.}'' Once again, the models’ consistent success across all formulations may be attributed to the presence of an explicit mathematical formulation in every version of the problem description.

\subsubsection{Vessel Loading}
involves positioning cargo boxes on a vessel such that all boxes fit within the available space and each is placed according to some distancing rules. 

We applied a context modification by reframing the task as arranging paintings in an art gallery, where each painting belongs to a specific category and must satisfy category-dependent distancing constraints. As in other cases, introducing only a context modification did not negatively affect the LLMs' performance. The only model that failed to produce a correct model was GPT-4, which assumed that the paintings were square and consequently ignored the second spatial dimension.

The distractor introduced in this problem stated: ``\textit{Curators want to maximise the aesthetic impact by placing the most visually striking artworks closest to the entrance and windows, while maintaining safety distances.}'' Both Claude 4 and DeepSeek-R1 added a ``reserved area'' constraint that was never mentioned in the description and ignored the explicit distractor. GPT-4, on the other hand, was influenced by the distractor and incorporated an additional constraint to maximise aesthetic impact.

\subsubsection{Perfect Square}
consists of packing a set of smaller squares, each with an integer side length, into a larger square such that no squares overlap and all sides are aligned with the borders of the enclosing square. In a perfect placement instance, all smaller squares have distinct sizes, and the sum of their areas equals the area of the enclosing square, meaning there is no unused space.

We applied a context modification by reframing the task as designing a park divided into distinct, non-overlapping sub-areas. Among the evaluated models, only Claude 4 failed to generate a correct model, introducing additional rules to prevent rectangular sub-areas. This appears to stem from a misinterpretation of the explanatory note indicating why all sub-squares were required to have different sizes.

To introduce a distractor, we added the following statement: ``\textit{We would like to minimise the unused areas by covering the whole park.}'' All three LLMs subsequently reformulated the task as an optimisation problem, thereby producing incorrect models.

\subsubsection{Social Golfers}
requires scheduling a set of golf games between a group of golfers such that no golfer plays in the same group as any other golfer on no more than one occasion. We aim to maximise the number of weeks of playing. 

We have introduced a context change, making the golf event into a research event where different researchers need to collaborate on different projects. Surprisingly, both GPT-4 and DeepSeek-R1 generated satisfaction models for both the original and context-modified problems. Besides that, GPT-4 also used a global constraint for symmetry breaking, but only in the modified problem, while it encoded the same constraint without using a global one for the original case. Both Claude 4 and GPT-4 used symmetry-breaking constraints in their models, while DeepSeek-R1 used them only in the original version.

When we introduced a distractor that seemingly required having interdisciplinary teams of scientists ("\textit{Scientists have different research areas (8 scientists each in areas A, B, C, and D). The coordinator wants each team to be as interdisciplinary as possible, ideally having one scientist from each area.}"), all three LLMs introduced a new constraint to model interdisciplinarity, even though it was never stated in the problem's objective.




\section{Discussion}
\label{sec:discussion}

Our study highlights the promises and pitfalls of using LLMs for CP modelling. 
While LLMs show strong syntactic and structural understanding when faced with familiar problems, their reasoning remains fragile and highly sensitive to how the problem is described. 
Through the systematic modification of well-known CP problems, we observed several consistent trends.

\paragraph{Words matter.}
The results clearly show that small textual variations can have disproportionate effects on the quality of generated models. 
Even the introduction of a single keyword, such as “maximise,” can completely shift the interpretation of the problem, leading to an incorrect optimisation formulation instead of a satisfaction one. 
This sensitivity underscores the importance of precise language and clear structural cues in prompts. 
Practitioners should treat the prompt not merely as natural language but as a specification, where lexical choices can act as hard constraints guiding the model’s reasoning path.

\paragraph{Contextual understanding remains shallow.}
Changing the surface context of a problem, e.g., reformulating car sequencing as knapsack assembly or Sudoku as a chess placement problem, can lead to incorrect or incomplete models. 
Although the underlying combinatorial structure is preserved, LLMs sometimes fail to abstract away from superficial narrative elements. 
This suggests that, when the combinatorial problem is vastly represented in the LLM training data, the LLM internal representations are largely tied to memorised textual patterns rather than true conceptual understanding of constraints and relations. 

\paragraph{Data contamination is a real concern.}
Our qualitative analysis of problem completion indicates that LLMs may rely heavily on memorised formulations of well-known benchmarks. 
Hence, benchmark results based solely on canonical formulations may overestimate the actual reasoning capabilities of LLMs in unseen settings.

\paragraph{Mathematical formulations anchor reasoning.}
When the description includes explicit algebraic or symbolic structure, LLMs are much less prone to distraction. 
Problems such as \textit{All-Interval Series} or \textit{Quasi Group Existence} consistently yielded correct models even under heavy rewording. 
This observation reinforces the idea that providing LLMs with clear mathematical anchors can significantly improve robustness and reduce the risk of misinterpretation.

\paragraph{Syntactical correctness of the model formulation}
We also observe a decline in the ability of the LLMs to produce syntactically correct and executable models when the problem description is altered. While some of these errors could, in principle, be resolved through iterative refinement (i.e., allowing the LLM to revise its own output), the observed decrease in performance suggests an over-reliance on memorised modelling patterns and, consequently, a limited degree of generalisation. Across all models and all LLMs, the only global constraint consistently employed was \textit{all\_different}, which is also the most common in the literature. This observation further supports the hypothesis that the LLMs rely primarily on familiar and frequently occurring modelling templates rather than abstract reasoning over the problem structure.

\subsection{Threats to Validity}

As with any empirical study, our analysis is subject to several threats to validity. 
We conclude this section by discuss these threats in terms of construct, internal, and external validity.

\paragraph{Construct validity.}
In our study, we assessed the modelling capabilities of LLMs through qualitative and quantitative metrics: \textit{run} and \textit{correct}.
Although these metrics provide interpretable signals of performance, they inevitably simplify the multifaceted notion of ``modelling quality.'' 
Similarly, the modified problem descriptions were manually designed to preserve the underlying combinatorial structure, yet we cannot guarantee that all reformulations are semantically equivalent or equally difficult. As a result, some of the observed differences in performance might reflect linguistic ambiguity rather than genuine differences in reasoning ability.

\paragraph{Internal validity.}
Since we lack access to the LLMs’ training data, we could only estimate data contamination indirectly, thus not being able to reach definitive conclusions about memorisation versus reasoning.

\paragraph{External validity.}
Our experiments focused on a limited set of ten problems from CSPLib plus one illustrative example. 
While this selection captures a broad range of modelling patterns, it does not represent the full diversity of real-world constraint programming applications. 
Different modelling languages, problem domains, or LLM architectures may yield different outcomes. 
Moreover, the use of English-only descriptions may bias performance against models trained on multilingual data. \label{sec:validity}

\section{Conclusions}\label{sec:conclusions}
We conducted a qualitative analysis of CP models generated by three different LLMs to evaluate their ability to generalise beyond previously seen problem descriptions. To this end, we introduced a systematic procedure for modifying existing problem statements so as to reduce the likelihood that an LLM could directly recall or reproduce a memorised model. We then compared the models produced from the original descriptions with those generated from the modified ones.

Our results indicate that context-based modifications generally reduce model quality, although the resulting models often remain semantically aligned with the original problem. A similar trend is observed when the original problem is less widely known, suggesting that familiarity plays a substantial role in successful modelling. In contrast, introducing distracting or irrelevant information typically leads to a marked decline in correctness: the LLMs frequently misinterpret the problem objective or prioritise secondary details, resulting in incorrect CP models. We find that LLMs struggle to recognise structural similarities across distinct but related problems, which limits their ability to produce robust or generalisable formulations.

We are interested in several directions for future work that could deepen our understanding of LLMs in constraint modelling. 
We are interested in developing interactive and in-context learning frameworks that allow LLMs to iteratively refine their formulations under controlled feedback while preserving interpretability. Another promising line concerns prompt-engineering strategies that integrate formal structure within natural language, helping the models to capture abstract relations rather than surface patterns. 
Finally, we aim to explore hybrid pipelines in which LLMs collaborate with domain-specific solvers, where the LLM proposes candidate models and the solver verifies, corrects, or guides refinement, thus combining the linguistic flexibility of LLMs with the rigour of formal reasoning systems.

\bibliography{bibliography}

\end{document}